\documentclass[twocolumn]{asme2e}
\usepackage[utf8x]{inputenc}

\usepackage{verbatim}
\usepackage{graphicx}
\usepackage{float}
\usepackage{caption}
\usepackage{amssymb} 
\usepackage{amsbsy}
\usepackage{amsmath}
\confshortname{Heywhale National Underwater Robot Professional Contest-Optics}

\confdate{Aril 30- May 27}
\confyear{2021}
\confcity{Zhan Jiang}
\confcountry{China}

\title{CDNet is all you need\\Cascade DCN Based Underwater Object Detection RCNN}

\author{Leader \& Member :Di Chang
    \affiliation{
	School of Information and Communication Engineering\\
	Dalian University of Technology\\
	Dalian, Liaoning\\
    Email: 2862588711@mail.dlut.edu.cn
    }	
}

\begin{document}

\maketitle    

\begin{abstract}
Object detection is a very important basic research direction in the field of computer vision and a basic method for other advanced tasks in the field of computer vision. It has been widely used in practical applications such as object tracking, video behavior recognition and underwater object detection. The Cascade-RCNN[1] and Deformable Convolution Network[2] are both classical and excellent object detection algorithms, so it is of practical significance to apply these two algorithms to underwater image object detection.
In this paper, the Cascade-RCNN and Deformable Convolution Network are used to detect the same underwater object dataset, and the performance of accuracy and detection speed of the two detection algorithms were explored under images of different sizes. In the data set used in this paper, the testing speed of the  Cascade-DCN network model reaches 2.2 seconds per task with Nvidia RTX 2080Ti GPU. In terms of accuracy, the accuracy of the  model is  0.47 with map50:95 method.And we set this as baseline model.
In this paper, better detection models are proposed for these two network models and this model has been renamed as CDNet. For the Cascade-RCNN model, the ResNext101[3] residual network is used to replace the ResNet50[4] residual network.Also Global Context Pooling[5] Block and Attention[6] Block is added at the end of the feature extraction stage. In addition to that we also implemented some other training tricks to the Neural Network.In the final stage of the B list we achieved $ 16^{th} $ place out of more than 500 teams in the accuracy part and $ 18^{th}  $ in the speed part.And our final ranking in this challenge is $ 18^{th}/508  $.\\Code is available at: \textit{https://github.com/Boese0601/2021-National-Underwater-Robotics-Vision-Optics}

\end{abstract}

\section{Introduction}

Underwater object detection is a branch of object detection research. Underwater is rich in mineral and biological resources, waiting for human to explore at the same time, there are many unknown dangers. In order to explore more safely and comprehensively, the research of underwater object detection and underwater robot has been put on the agenda. The underwater robot is designed to replace human beings to go deep into the seabed and complete the underwater scientific and military mission investigation, marine resource assessment, seabed geological testing and other dangerous tasks. Underwater object detection is the key basic technology to help the underwater robot to complete these underwater tasks better.
However, the underwater environment is very different from the land environment, which makes the object recognition more difficult. Due to the absorption and scattering characteristics of light in the water and other reasons, there will be some problems affecting the image quality, such as color distortion, blurring, contrast distortion and so on, when there is not enough light source in the water, resulting in the low accuracy of underwater image object detection. In the reality that the recognition accuracy of underwater object detection algorithm is still relatively low, the research of underwater object detection algorithm is very necessary.

\section{Related Work}
In this part we introduce our baseline model with standard Cascade-RCNN and its deformable convolution operation,as well as the Feature Pyramid Network[7] neck.
\subsection{Cascade RCNN}
While the ideas proposed in this work can be applied to
various detector architectures, we focus on the popular two-stage architecture of the Faster R-CNN, shown in Fig. 1
(a). The first stage is a proposal sub-network, in which the
entire image is processed by a backbone network, e.g. ResNet
[27], and a proposal head (“H0”) is applied to produce
preliminary detection hypotheses, known as object proposals. In the second stage, these hypotheses are processed by
a region-of-interest detection sub-network (“H1”), denoted
as a detection head. A final classification score (“C”) and a
bounding box (“B”) are assigned per hypothesis. The entire
detector is learned end-to-end, using a multi-task loss with
bounding box regression and classification components.
\\
\quad In order to generate high quality detection, we use Cascade-RCNN. Following the original implementation, we set the IoU thresholds to 0.5, 0.6 and 0.7 for each RCNN stage respectively. We also try different IoU thresholds, and find that the default setting yields best
performance.
\begin{center}
	\begin{figure}[H]
		\centering                                     
		\includegraphics[width=0.5\textwidth]{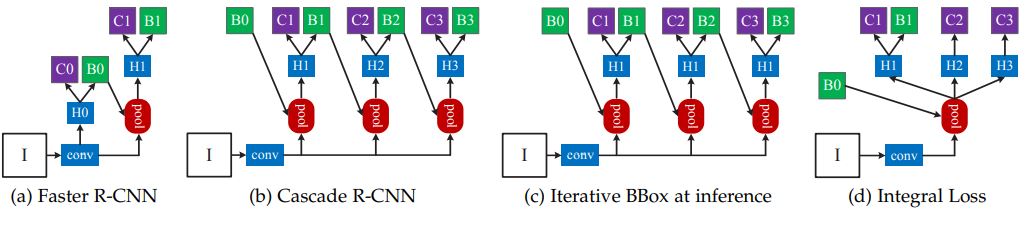}   
		\caption{Cascade-RCNN}                                 		\label{img}                                    
	\end{figure}
\end{center}

\subsubsection{bounding box regression}
\quad\\
A bounding box $\mathbf{b}=\left(b_{x}, b_{y}, b_{w}, b_{h}\right)$ contains the four coordinates of an image patch $\mathrm{x}$. Bounding box regression aims to regress a candidate bounding box $\mathbf{b}$ into a target bounding box $\mathbf{g}$, using a regressor $f(\mathbf{x}, \mathbf{b})$. This is learned from a training set $\left(\mathbf{g}_{i}, \mathbf{b}_{i}\right)$, by minimizing the risk
$$
\mathcal{R}_{\text{loc }}[f]=\sum_{i} L_{\text{loc }}\left(f\left(\mathbf{x}_{i}, \mathbf{b}_{i}\right), \mathbf{g}_{i}\right)
$$
As in Fast R-CNN [21],
$$
L_{l o c}(\mathbf{a}, \mathbf{b})=\sum_{i \in\{x, y, w, h\}} \text{ smooth }_{L_{1}}\left(a_{i}-b_{i}\right)
$$
where
$$
\text{ smooth }_{L_{1}}(x)=\left\{\begin{array}{cl}
	0.5 x^{2}, & |x|<1 \\
	|x|-0.5, & \text { otherwise }
\end{array}\right.
$$
is the smooth $L_{1}$ loss function. To encourage invariance to scale and location, smooth $_{L_{1}}$ operates on the distance vector $\Delta=\left(\delta_{x}, \delta_{y}, \delta_{w}, \delta_{h}\right)$ defined by
$$
\begin{array}{c}
	\delta_{x}=\left(g_{x}-b_{x}\right) / b_{w}, \quad \delta_{y}=\left(g_{y}-b_{y}\right) / b_{h} \\
	\delta_{w}=\log \left(g_{w} / b_{w}\right), \quad \delta_{h}=\log \left(g_{h} / b_{h}\right)
\end{array}
$$

\subsection{Deformable Convolution Network}
The DCN consists of two parts of operation.The first is deformable convolution. It adds 2D offsets to the regular grid sampling locations in the standard convolution. It enables free form deformation of the
sampling grid. It is illustrated in Figure 2. The offsets
are learned from the preceding feature maps, via additional
convolutional layers. Thus, the deformation is conditioned
on the input features in a local, dense, and adaptive manner.
The second is deformable RoI pooling. It adds an offset
to each bin position in the regular bin partition of the previous RoI pooling. Similarly, the offsets are learned from the preceding feature maps and the RoIs, enabling
adaptive part localization for objects with different shapes.
Both modules are light weight. They add small amount of parameters and computation for the offset learning. They can readily replace their plain counterparts in deep CNNs and can be easily trained end-to-end with standard backpropagation. The resulting CNNs are called deformable
convolutional networks, or deformable ConvNets.
\begin{center}
	\begin{figure}[H]
		\centering                                     
		\includegraphics[width=0.5\textwidth]{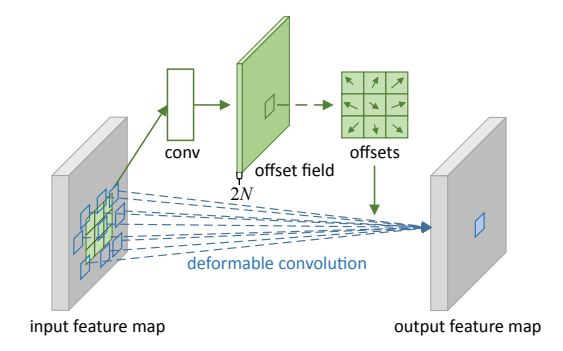}   
		\caption{Deformable Convolution }                                  %
		\label{img}                                    
	\end{figure}
\end{center}
\begin{center}
	\begin{figure}[H]
		\centering                                      
		\includegraphics[width=0.5\textwidth]{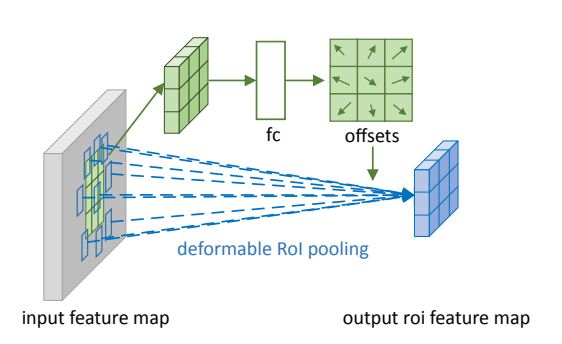}   
		\caption{Deformable RoIPooling Network}                               
		\label{img}                                    
	\end{figure}
\end{center}
\subsection{Feature Pyramid Network}
Our method takes a single-scale image of an arbitrary
size as input, and outputs proportionally sized feature maps
at multiple levels, in a fully convolutional fashion. This process is independent of the backbone convolutional architectures, and in this paper we present results using ResNext. The construction of our pyramid involves a bottom-up pathway, a top-down pathway, and lateral connections, as introduced in the following. Bottom-up pathway. The bottom-up pathway is the feedforward computation of the backbone ConvNet, which computes a feature hierarchy consisting of feature maps at several scales with a scaling step of 2. There are often many layers producing output maps of the same size and we say these layers are in the same network stage. For our feature pyramid, we define one pyramid level for each stage. We choose the output of the last layer of each stage as our reference set of feature maps, which we will enrich to create our pyramid. This choice is natural since the deepest layer of each stage should have the strongest features.Specifically, for ResNext we use the feature activations output by each stage’s last residual block. We denote the output of these last residual blocks as {C2, C3, C4, C5} for conv2, conv3, conv4, and conv5 outputs, and note that they have strides of {4, 8, 16, 32} pixels with respect to the input image. We do not include conv1 into the pyramid due to its large memory footprint.The structure can be displayed as Fig.4

\begin{center}
	\begin{figure}[H]
		\centering                                      
		\includegraphics[width=0.5\textwidth]{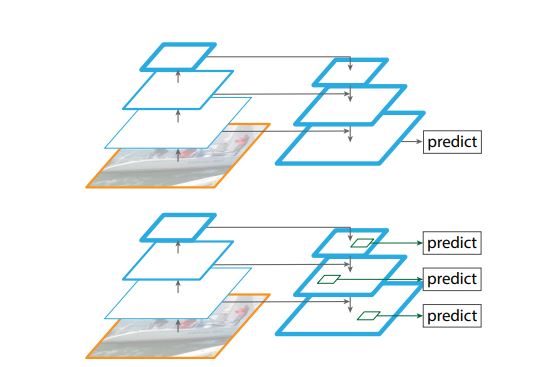}   
		\caption{Feature Pyramid Network}                                 
		\label{img}                                    
	\end{figure}
\end{center}
\section{Approach}
\subsection{Network Structure}
The whole network can be divided into two parts as the traditional two-stage detection algorithm.The first part is the feature extraction backbone and the second part is feature fusion neck and detection predition head.
\subsubsection{Backbone}\quad \\
In this case we apply ResNext101 as our powerful backbone,because ResNext101 combines the advantages of both ResNet and InceptionNet,which maintains residual connection between convolution blocks and also have a wide forward block like InceptionNet.At the same time we pretrained this part of the network on coco dataset and freeze the  parameters of the first convolution block and trained the other parts with  our underwater dataset end-to-end.
ResNeXt101 Backbone model could be seen as Fig.5.\\
\begin{center}
	\begin{figure}[H]
		\centering                                     
		\includegraphics[width=0.5\textwidth]{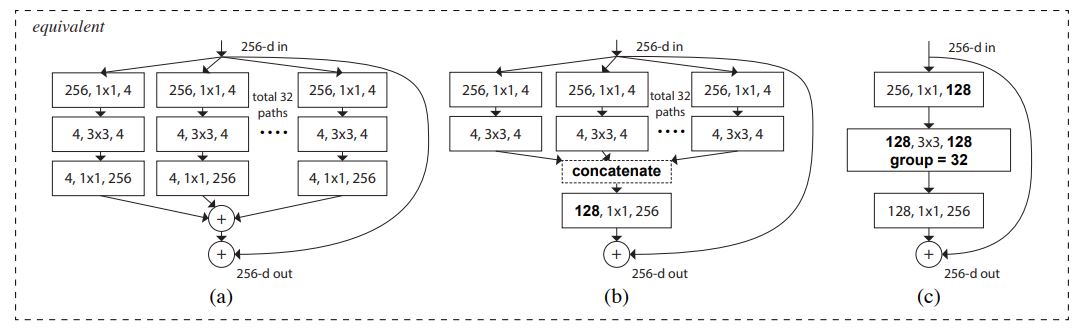}   
		\caption{. Equivalent building blocks of ResNeXt. (a): Aggregated residual transformations right. (b): A block equivalent to (a), implemented as early concatenation. (c): A block equivalent to (a,b), implemented as grouped convolutions. Notations in bold
			text highlight the reformulation changes. A layer is denoted as input channels, filter size, output channels}                                 
		\label{img}                                    
	\end{figure}
\end{center}
Except from the basic feature extraction backbone model,we also applied some plugins to this part.First we use non-local structure to replace the traditional average pooling and added global context information to our extracted feature map.We called this part the \textit{gcb plugins}.\\
After the basic backbone and the gcb plugins,the feature map is positional-encoded and fed into another structure called attention block.This kind of structure is totally the same as the structure being mentioned in the paper attention is all you need.We cannot use the traditional qkv+postional encoding because of the limitness of GPU resource,so we just use single-head attention with positional encoding instead of the multi-head attention.In this way we renamed this structure called attention plugins.
\subsubsection{Neck}\quad \\
Traditional FPN does not have a feature fusion operation after the backbone layers.We rethink the FPN structure according to the paper YOLOF you only look on one-level feature.In this paper the author mentioned that the effectness of FPN does not come from the multi-level faeture structure but the encoder=decoder structure ,for implementation details of Feature Pyramid Neural Network please infer to the first section of this paper.\\
Instead of the traditional FPN we apply BFP struture as the Libra RCNN and also we extract the advantage of NASFPN[8],this structure is shown in Fig.6.For the encoder- decoder structure is really an important idea in the Feature Pyramid Network,so we just simply do a fusion operation at the end of the Multi-level feature and encoded these feature maps into an single-level feature map.The we did a sequence of up-sampling operation and reconstruct the structure as U-Net,then we predict the anchor on each level of the feature maps as Cascade RCNN.

\begin{center}
	\begin{figure}[H]
		\centering                                      
		\includegraphics[width=0.5\textwidth]{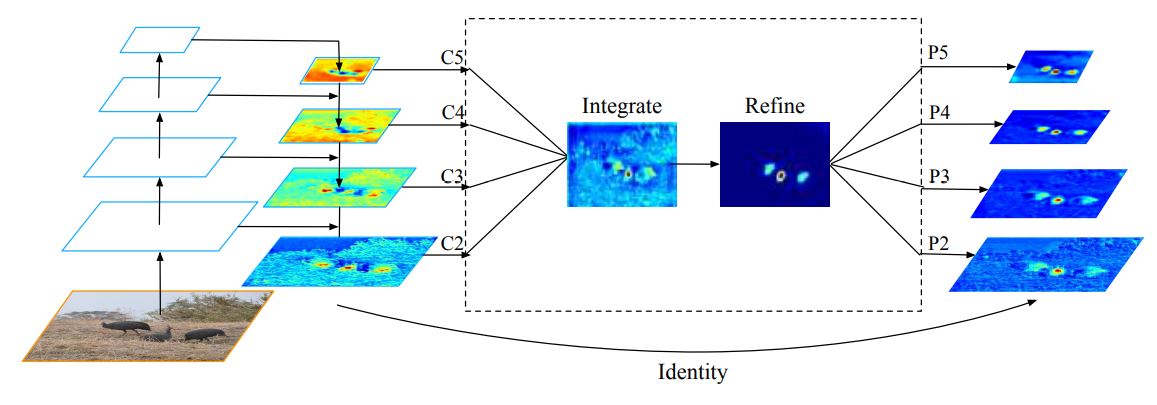}   
		\caption{NAS-FPN}                               
		\label{img}                                    
	\end{figure}
\end{center}
\subsubsection{Detection Head}\quad \\
The Structure here is basically the same as original Cascade-RCNN,we also tried some other types of loss function,such as GIoU Loss,CIoU Loss and DIoU Loss.But none of them perform as well as the simple SmoothL1 Loss function.Also,in terms of the type of the sampler,we tried InstanceBalance Sampler which was introduced in Libra-RCNN and solved the problem of disparity between different classes of objects.Our dataset consists of  pictures of four different types of underwater creatures,holothurian,echinus,scallop and starfish.In our case the amount of scallop is quite low and we did a  simple count that we fould the number of the starfish is ten times of that of the starfish.So the InstanceBalance Sampler should have solved this kind of problem,but to my surprise it have  not.And GIoU loss have the same problem.In the orginal paper GIoU performed quite well on COCO Dataset,but in our case GIoU and its improved versions CIoU and DIoU all lost their effect.So we just used the original Cascade detection head and applied soft-nms at the end of the algorithm instead of the simple nms.The detection head Structure is shown as following section.
\\
The initial hypotheses distribution produced by the RPN is heavily tilted towards low quality. For example, only 2.9\% of examples are positive for an IoU threshold u = 0.7. This makes it difficult to
train a high quality detector. The Cascade R-CNN addresses
the problem by using cascade regression as a resampling
mechanism. This is inspired by Faster-RCNN, where nearly
all curves are above the diagonal gray line, showing that
a bounding box regressor trained for a certain u tends
to produce bounding boxes of higher IoU. Hence, starting
from examples, cascade regression successively resamples an example distribution of higher IoU.
This enables the sets of positive examples of the successive
stages to keep a roughly constant size, even when the detector quality u is increased. Figure 4 illustrates this property,
showing how the example distribution tilts more heavily
towards high quality examples after each resampling step.\\
At each stage t, the R-CNN head includes a classifier
$ h_{t} $ and a regressor $ f_{t} $ optimized for the corresponding IoU threshold $ u_{t}$, where $  u_{t} > u_{t−1} $ . These are learned with loss:\\
\quad \\
\begin{center}
	$ L\left(\mathbf{x}^{t}, g\right)=L_{c l s}\left(h_{t}\left(\mathbf{x}^{t}\right), y^{t}\right)+\lambda\left[y^{t} \geq 1\right] L_{l o c}\left(f_{t}\left(\mathbf{x}^{t}, \mathbf{b}^{t}\right), \mathbf{g}\right) $\\
\end{center}
\quad \\
where $ b
t = f_{t−1} (x_{
t−1}, b_{t−1}) $, g is the ground truth object for
$ x_{t} $
, $\lambda$ = 1 the trade-off coefficient, $ y_{t} $ is the label of $ x_{t} $ under the $u_{t}$ criterion, [·] is the indicator function.
Note that the use of [·] implies that the IoU threshold u
of bounding box regression is identical to that used for
classification. This cascade learning has three important
consequences for detector training. First, the potential for
overfitting at large IoU thresholds u is reduced, since positive examples become plentiful at all stages.Second, detectors of deeper stages are optimal for higher IoU thresholds. Third, because some outliers are removed as the IoU threshold increases,the learning
effectiveness of bounding box regression increases in the
later stages. This simultaneous improvement of hypotheses
and detector quality enables the Cascade R-CNN to beat
the paradox of high quality detection. At inference, the
same cascade is applied. The quality of the hypotheses
is improved sequentially, and higher quality detectors are
only required to operate on higher quality hypotheses, for
which they are optimal. 

\section{Training Policy}
\subsection{Hyperparameter}
The entire model is trained with 4 Nvidia RTX 3090 GPU end2end with memory 64GB.We used SGD as optimizer and set the learning rate to 0.005(0.00125*number of gpus).The learning rate would drop on epoch 8 and 11 to 0.0005 and 0.0001 and have a warmup process at the beginning of epoch 1 with 500 iterations,this training policy has been proved to be effective with most computer vision tasks.We set the soft-nms threshold to 0.7 with rpn and 0.5 with rcnn so that the anchors would be filtered with suitable constraints.\\
We also changed the default confidence threshold from 0.3 to 0.0001,in this way the integration value of recall and accuracy will not lose too much points.
\subsection{Network Architecture}
We tried some ways to improve our ResNext backbone,as introduced in the previous sections.The context block gcb and attention block perform quite well with a huge improvement in map value.DCN is the part of the baseline model,and we must admit that this part is the key of many effective tricks.We also tried BFP[9] in the Feature Pyramid Network,but it seems not so useful on our dataset as the orginal coco dataset experiment.And GIoU[10] loss,CIoU[11] loss,DIoU[11] loss all did not work at all which is an astonishing phenomenon.The experiment result could be seen in the next section.
\\

\subsection{Data Augumentation Tricks}
We also implemented several training tricks to augument our pictures.Including RandomRotate 90 degrees,Random Flipping,Vertical Flipping,Cutout,Mixup and Multi-scale training and testing.At the final stage of B list.The only trick that we kept is RandomRotate,while the others will lead to worse robustness of the model.

\section{Result on Underwater Dataset}
In this section the testing result on the A list will be presented in
the form and we will see the final result on the B list.Note only the essential part of the evaluation is given.Some of the results has been removed because the adjustment of simple hyperparameters is not so intereseting.

\begin{table}[H]
	\resizebox{260pt}{100pt}{
	\begin{tabular}{|l|l|}
		\hline
		model                                                             & map@50:95                      \\ \hline
		baseline:Cascade+FPN+DCN+X101+Random90+Multi-scale                & 0.523  \\ \hline
		baseline without dcn                                              & 0.527  \\ \hline
		baseline+dcn pretrained on coco                                   & 0.549  \\ \hline
		baseline+dcn pretrained+gcb                                       & 0.561  \\ \hline
		baseline+dcn pretrained+gcb+data cleansing                        & 0.563512  \\ \hline
		baseline+dcn pretrained+gcb+data cleansing+GIoU                   & 0.553  \\ \hline
		baseline+dcn pretrained+gcb+data cleansing+CIoU                   & 0.554  \\ \hline
		baseline+dcn pretrained+gcb+data cleansing+Cutout                 & 0.556  \\ \hline
		baseline+dcn pretrained+gcb+data cleansing+MixUp                  & 0.557  \\ \hline
		baseline+dcn pretrained+gcb+data cleansing+InstanceBalanceSampler & 0.5604 \\ \hline
		baseline+dcn pretrained+gcb+data cleansing+BFP                    & 0.5604 \\ \hline
		baseline+dcn pretrained+gcb+data cleansing+attention              & 0.563542  \\ \hline
		baseline+dcn pretrained+gcb+data cleansing+BBoxJitter             & 0.563662       \\\hline
		baseline+dcn pretrained+gcb+data cleansing+attention+BBoxJitter   & 0.567  \\ \hline
	
	\end{tabular}
}
\caption{Testing Result}
\end{table}
It is noteworthy that we add a new type of data augumentation called Bounding Box Jitter-BBoxJitter.This trick aims to adjust the bounding box of ground truth labels.Because the given training dataset contains Labeling noise which means that some of the locations of ground truth boxes are given by mistake on purpose.This trick only performs well in this typical circumstances.The offcial datasets would not have such labeling noises so do NOT try this tricks on them!!!
\\
\quad \quad \\Our test result on B list will come soon,please refer to the ofiicial heywhale website.
\section{Conclusion}
In this paper we present a new model called CDNet in the field of underwater detection challenge.This  model is not perfect but effective in this kind of environment.We achieved acceptable result though there's still a long way to go until the state-of-art algorithm.According to my conprehension of this task,tiny object and overlapping object detection is the most vital direction of future works with wide range of improvement.At last thanks to supervisor Prof.Dr.  Dong Wang and Prof.Dr. Yifan Wang.They kindly gave me efficient support with GPU resource and I'm really grateful for other kinds of contribution.

\end{document}